\documentclass{article}
\usepackage{amsmath,epsfig}
\usepackage[preprint]{spconfa4}
\usepackage{xcolor}
\usepackage{cite}

\usepackage{amssymb,bm}
\usepackage{bbding}
\usepackage{booktabs}
\usepackage{bbding} 
\usepackage{multirow}
\usepackage[normalem]{ulem}
\useunder{\uline}{\ul}{}

\let\OLDthebibliography\thebibliography
\renewcommand\thebibliography[1]{
  \OLDthebibliography{#1}
  \setlength{\parskip}{0pt}
  \setlength{\itemsep}{0pt plus 0.3ex}
}

\begin{document}\sloppy

\def\x{{\mathbf x}}
\def\L{{\cal L}}

\title{Learning Appearance-motion Normality for Video Anomaly Detection}
%
\name{Yang Liu\textsuperscript{1}, Jing Liu\textsuperscript{1}, Mengyang Zhao\textsuperscript{1}, Dingkang Yang\textsuperscript{1}, Xiaoguang Zhu\textsuperscript{2}, Liang Song\textsuperscript{1}$^\ast$\thanks{$^\ast$Corresponding author. This work is supported by the Shanghai Key Research Laboratory of NSAI.}}
\address{\textsuperscript{1}Academy for Engineering \& Technology, Fudan University, Shanghai, China\\\textsuperscript{2}SEIEE, Shanghai Jiao Tong University, Shanghai, China
}

\maketitle

\begin{abstract}
  Video anomaly detection is a challenging task in the computer vision community. Most single task-based methods do not consider the independence of unique spatial and temporal patterns, while two-stream structures lack the exploration of the correlations. In this paper, we propose spatial-temporal memories augmented two-stream auto-encoder framework, which learns the appearance normality and motion normality independently and explores the correlations via adversarial learning. Specifically, we first design two proxy tasks to train the two-stream structure to extract appearance and motion features in isolation. Then, the prototypical features are recorded in the corresponding spatial and temporal memory pools. Finally, the encoding-decoding network performs adversarial learning with the discriminator to explore the correlations between spatial and temporal patterns. Experimental results show that our framework outperforms the state-of-the-art methods, achieving AUCs of 98.1\% and  89.8\% on UCSD Ped2 and CUHK Avenue datasets.
\end{abstract}
\begin{figure*}
  \centering
  \includegraphics[width=\textwidth]{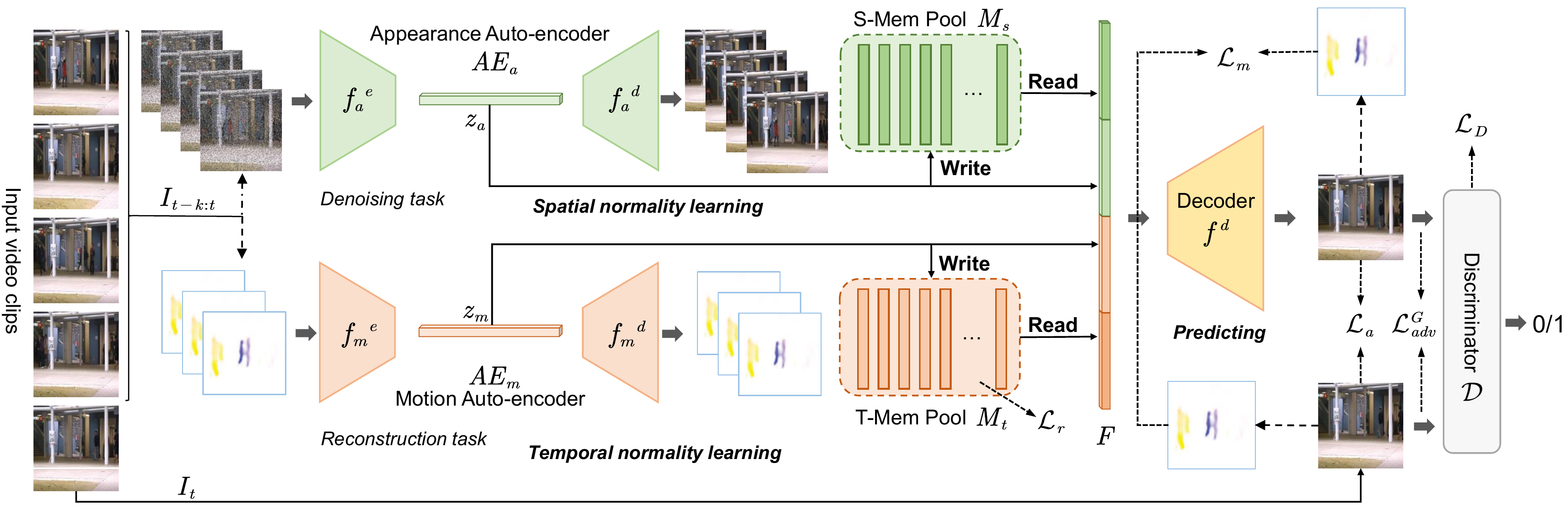}
  \caption{Overview of the proposed saptial-temporal memories augmented two-stream auto-encoder framework.}
  \label{fig:1}
\end{figure*}
\section{Introduction}
Video anomaly detection (VAD) is the key technology of intelligent surveillance systems, which aims to automatically detect and locate abnormal events from videos \cite{9739763}. Due to the ambiguity of the definition of anomaly, the label of an event depends on the scenario. For example, fast running indoor is considered abnormal, while outdoor one is usually defined as normal. Therefore, VAD is always formulated as an out-of-distribution detection task, which learns a model of normality that can well describe normal events. Events that cannot be described are discriminated against as abnormal. Additionally, most existing VAD methods \cite{luo2017revisit,liu2018future,gong2019memorizing,9739751} are unsupervised due to the difficulty of collecting abnormal events and obtaining precise frame-level labels. Earlier methods, such as one-class support vector machine, use normal samples to train a one-class classifier that attempts to find a hyperplane as the boundary of normal events, while events whose features fall outside the boundaries are identified as anomalies. Such methods will suffer from the curse of dimensionality. 

Recently, motivated by the success of deep learning in video understanding, various unsupervised deep VAD methods \cite{zhang2020normality,cai2021appearance,chang2020clustering} have been proposed. They use only normal videos to train deep convolutional neural networks (CNNs) to solve a proxy task, such as reconstructing input video clips \cite{luo2017revisit,gong2019memorizing} or predicting future frames \cite{liu2018future,ye2019anopcn}. The assumption is that models learned on normal events cannot compress abnormal events that have not been seen before, leading to large reconstruction or prediction errors. Therefore, anomalies can be discriminated by measuring the deviation between the testing samples and the learned model.

Video anomalies typically include appearance and motion anomalies, corresponding to semantic shifts of information in the spatial and temporal dimensions, respectively. Thus, the key to unsupervised VAD is to learn the prototypical appearance and motion patterns of normal events. However, most existing methods \cite{gong2019memorizing,park2020learning,liu2018future} do not consider the difference of spatial and temporal features in isolation. In contrast, they train deep CNNs to learn spatial-temporal features by performing a single proxy task, lacking the exploration of unique appearance and motion patterns. Although some methods adopt reconstruction and prediction tasks simultaneously \cite{zhao2017spatio}, they still do not treat appearance and motion anomalies differently. In recent years, two-stream structures \cite{chang2020clustering,doshi2020continual,yu2020cloze,cai2021appearance} have been proposed to separately learn appearance and motion patterns. For example, Chang \textit{et al.} \cite{chang2020clustering} trained two auto-encoders to obtain appearance and motion codings and used clustering algorithms to determine the boundaries of normal events. Li \textit{et al.} \cite{li2020spatial} proposed cascaded auto-encoders to perform frames reconstruction and optical flow reconstruction tasks separately. Then, the appearance anomaly score and motion anomaly score are calculated independently. Most two-stream structures lack the exploration of the correlation between appearance and motion features.

In this paper, we propose spatial-temporal memories augmented two-stream auto-encoder (STM-AE) framework, which learns appearance and motion normality in isolation and explores the correlations via a prediction task. In the training phase, the appearance and motion features are extracted by two parallel auto-encoders, which are pre-trained by the denoising and optical flow generation proxy tasks. Then, the prototypical patterns are written into the corresponding memory pools through a top-$k$ attention-based updating mechanism. Finally, we retrieve memory items and feed the aggregated features to the decoder to predict future frames. To improve the predicting ability of the decoder, we introduce a patch discriminator \cite{mao2017least} for adversarial learning. In the testing phase, the spatial and temporal features of testing videos are rewritten by the learned memory pools so that the prediction results are close to normal videos. Therefore, abnormal events will lead to large prediction errors. The main contributions are summarised as follows: 
\begin{itemize}
  \item We propose spatial-temporal memories augmented auto-encoder framework to learn and record prototypical appearance and motion patterns independently. Besides, we introduce adversarial learning to explore the correlations between appearance and motion normality.
  \item To record the prototypical patterns, we propose top-$k$ attention to update the memory pools. Only part of the items is rewritten in each update step to ignore the personalized patterns of normal events.
  \item Experimental results demonstrate that the proposed framework outperforms the state-of-the-art method with AUCs of 98.1\%, 89.8\%, and 73.8\% on UCSD Ped2, CUHK Avenue, and ShanghaiTech, respectively.
\end{itemize}
\section{Method}
\subsection{Architecture}
As shown in Figure \ref{fig:1}, the proposed STM-AE framework consists of a two-stream structure, where the appearance auto-encoder $AE_a$ is used for learning spatial features and motion auto-encoder $AE_m$ for temporal features. In the training phase, we first use normal vides to pre-train $AE_a$ and $AE_m$. At time $t$, the $AE_a$ compresses the input RGB images $\bm{I}_{t-k:t}$ into low-dimensional appearance features $\bm{z}_a$ by performing a denoising task, where $k$ denote the temporal length of input clips. In contrast, the $AE_m$ obtain the motion features $\bm{z}_m$ by performing an optical flow reconstruction task. Unlike other two-stream structures, we replace the images reconstruction with the denoising task for $\bm{z}_a$. In contrast, the denoising task requires the encoder to fully understand the spatial information since $AE_m$ needs to reason out the missing parts (noise-masked regions) from the given parts (noise-free regions). The optical flow frames can reflect the motion information while ignoring background, so that we use the optical flow frames reconstruction as the proxy task of $AE_m$. The input optical flow frames are calculated by the Flownet \cite{dosovitskiy2015flownet}. 

The two auto-encoders share the same structure, consisting of a 5-layer encoder and a 6-layer decoder. Each layer is a $3\times 3$ convolutional layer followed by batch normalization and non-linear activation. To ensure the diversity of $\bm{z}_a$ and $\bm{z}_m$, we use the LeakyReLU as a non-linear activation layer of the encoder to retain negative values. In the pre-training phase, the goal of $AE_a$ and $AE_m$ is to learn how to compress appearance and motion patterns. We set the objective function for the denoising task to be minimizing the $L_2$ distance between the generated frames and ground truth, as follows:
\begin{equation}
  \label{equ:1}
  \underset{\theta_a^e, \theta_a^d}{\min} \parallel f_a^d\left(f_a^e\left(\tilde{\bm{I}}_{t-k:t}; \theta_a^e\right); \theta_a^d\right)-{\bm{I}}_{t-k:t}\parallel_2^2
\end{equation}
where $\tilde{\bm{I}}_{t-k:t}$ donetes the video clips with 20\% salt and pepper noise. $\theta_a^e$ and $\theta_a^d$ denote the parameter of the appearance encoder $f_a^e$ and decoder $f_a^d$. Similarly, the objective function for the reconstruction task is to optimise motion encoder $f_m^e$ and decoder $f_m^d$ by minimising the $L_2$ distance between the reconstructed optical flow and input ones, as follows:
\begin{equation}
  \label{equ:2}
  \underset{\theta_m^e, \theta_m^d}{\min} \parallel f_m^d\left(f_m^e \left(f \left(\bm{I}_{t-k:t} \right); \theta_m^e \right); \theta_m^d \right)-f \left({\bm{I}}_{t-k:t} \right)\parallel_2^2
\end{equation}
where $f(\cdot)$ denote the Flownet. After pre-training, $\theta_a^d$ and $\theta_m^d$ are frozen. $\bm{z}_a$ and $\bm{z}_m$ are written to the corresponding memory pools as prototype patterns of normal events. In addition, $\bm{z}_a$ and $\bm{z}_m$ are used as queries to retrieve the relevant memory items. We aggregate the raw features and retrieved items and feed them to the decoder to predict future frames. Since the memory pool is updated via normal events only during the training phase, the aggregated features $\bm{F}$ will be close to the representation of normal events in the feature space. In the testing phase, the $\bm{F}$ of the abnormal event will deviate from the direct concatenation of its $\bm{z}_m$ and  $\bm{z}_a$, which in turn leads to large prediction errors. To improve the quality of the predicted frames, we measure the difference of the generated frames in pixels and optical flow to ensure the completeness of the appearance and motion information.
\subsection{Appearance and Motion Normality Recording}
The spatial and temporal memory pools are used to record the prototypical appearance and motion pattern of normal events, respectively. Both the two pools contains $N$ memory items of dimension $C$, denoted by the two-dimensional matrix $\bm{M}_s$ and $\bm{M}_t$. Since the two pools share the same updating mechanism, for simplify, only the learning process of spatial normality is presented here. The write operation aims to writing $\bm{z}_a$ to $\bm{M}_s$. The prototypical spatial pattern of normal events is recorded in $\bm{M}_s$ by updating the items in the training phase. Firstly, we flatten the $\bm{z}_a$ along the spatial dimension and obtain $\hat{N}$ query vectors of dimension $C$, where $C$ equals the number of channels of $\bm{z}_a$. The size of appearance features are $H\times W \times C$, so $\hat{N}=H\times W$. We treat the memory items as linear combinations of the queries. In each update step, the weighted queries are written to the history memory items as follows:
\begin{equation}
  \label{equ:3}
  \hat{\bm{m}}_s^i = g\left(\bm{m}_s^i + \bm{w}_s^i\bm{Q}\right)
\end{equation}
where $g(\cdot)$ denotes the $L_2$ normalization, which are used to keep the data scale of the updated memory items $\hat{\bm{m}}_s^i$ same as history memory items $\bm{m}_s^i$. $\bm{w}_s^i$ is a weight vector of size $1\times \hat{N}$, used to weight queries $\bm{Q}$. Different from previous work \cite{park2020learning}, considering the diversity of normal events, only part of the query vectors are written into the memory pool instead of all, which helps the pool to record general patterns of normal events. Therefore, we only retain the weight of the top-$k$ relevant query vectors when computing $\bm{w}_s^i$. Specifically, We first compute the cosine similarity of $\bm{m}_s^i$ to all queries, denoted by $\tilde{\bm{w}}^i$. Then, the top-$k$ values of $\tilde{\bm{w}}^i$ are retained, and the others are set to 0. We perform softmax normalization on $\tilde{\bm{w}}^i$ to calculate the weights $\bm{w}_s^i$, as follows:
\begin{equation}
  \bm{w}_s^i=\frac{\exp \left(\tilde{\bm{w}}^i\right)-1}{\sum_{j=1}^{\hat{N}} \exp \left(\tilde{\bm{w}}^i_j\right)-\hat{N}}
\end{equation}
The read operation is to reconstruct the queries $\bm{Q}$ using the memory items. We treat the reconstructed query $\hat{\bm{z}}_a$ as the linear combination of the memory items. The weights are calculated by performing softmax normalization on the cosine similarity of the query to all memory items.
\subsection{Adversarial Learning and Frame Prediction}
To explore the correlation between appearance and motion features, we concatenate the reconstructed and raw features and feed them to the 5-layer decoder $f^d$ for predicting future frames. We measure the quality of the predicted images at both the pixel domain and optical flow domain to ensure spatial and temporal information integrity. The appearance loss $\mathcal{L}_a$ is defined as the $L_2$ distance between the predicted frame $\hat{\bm{I}}_t$ and ground truth $\bm{I}_t$, as follows:
\begin{equation}
  \label{equ:5}
  \mathcal{L}_a = \parallel \hat{\bm{I}}_t - \bm{I}_t \parallel_2^2
\end{equation}
Similarity, we define motion loss $\mathcal{L}_m$ as follows:
\begin{equation}
  \label{equ:5}
  \mathcal{L}_m = \parallel f(\hat{\bm{I}}_t, \bm{I}_{t-1})-f\left(\bm{I}_t, \bm{I}_{t-1}\right) \parallel_2^2
\end{equation}
In addition, to further improve the ability to fuse and decode appearance and motion features, We treat the two encoders $f_a^e$ and $f_m^e$ and the decoder $f^d$ as the generator $\mathcal{G}$. A patch discriminator $\mathcal{D}$ \cite{mao2017least} is introduced to perform adversarial learning with $\mathcal{G}$. The discriminator accepts image patches as input and discriminates whether the input is from real future frames (output 0) or predicted ones (output 1). In contrast, the objective of $\mathcal{G}$ is to generate frames that can be discriminated against as real ones by the discriminator. The adversarial loss for training the generator, denoted by $\mathcal{L}_{adv}^G$, is as follows:
\begin{equation}
  \mathcal{L}_{adv}^G =\sum_{i, j} \frac{1}{2} \left(\mathcal{D}(\hat{\bm{I}}_t^{i,j})-1\right)^2
\end{equation} 
where $i,j$ are the spatial index of patches. The loss for training the discriminator, denoted by $\mathcal{L}_D$, is defined as follows:
\begin{equation}
  \mathcal{L}_D =\sum_{i, j} \frac{1}{2}\left(\mathcal{D}(\bm{I}_t^{i,j})- 1\right)^2 +\sum_{i, j} \frac{1}{2}\mathcal{D}(\hat{\bm{I}}_t^{i,j})^2
\end{equation}
Additionally, we introduce $\mathcal{L}_{r}$ to ensure the representativeness of the memory items, as follows:
\begin{equation}
  \mathcal{L}_{r} = \sum_{i=1}^{N}\parallel \bm{q}^i - \bm{m}^i_1 \parallel_2^2 - \parallel \bm{q}^i - \bm{m}^i_2 \parallel_2^2
\end{equation}
where $\bm{m}^i_2$ and $\bm{m}^i_2$ denote the first and second nearest memory items to query $\bm{q}^i$. Balanced by $\lambda_1$, $\lambda_2$ and $\lambda_3$, the total loss for training the generator, denoted by $\mathcal{L}_G$, is as follows:
\begin{equation}
  \label{equ:8}
  \mathcal{L}_G =  \mathcal{L}_a + \lambda_1 \mathcal{L}_m + \lambda_2 \mathcal{L}_{r} + \lambda_3 \mathcal{L}_{adv}^G
\end{equation}
\subsection{Anomaly Score}
After training, we assume that the prototypical appearance and motion features of normal events have been recorded in two memory pools. The generator can efficiently compress and predict normal events so that the prediction error is relatively low. In contrast, the aggregated features of abnormal events will be close to normal events, which leads to poor prediction results. Therefore, we obtain the anomaly score by calculating the prediction error $e$. Follow previous works \cite{liu2018future,park2020learning}, $e$ is defined as the peak signal-to-noise ratio between the predicted frames $\hat{\bm{I}}_t$ and the ground truth $\bm{I}_t$, as follows:
\begin{equation}
  \label{equ:6}
  e_t=10 \log _{10} \frac{255^2}{\parallel \hat{\bm{I}}_{t}-\bm{I}_{t}\parallel_{2}^{2}}
\end{equation}
Finally, we use maximum-minimum normalization to map $e_t$ to anomaly score $s_t$ in the range of [0, 1], as follows:
\begin{equation}
  s_t=\frac{e_t-\min_t e_t}{\max_t e_t-\min_t e_t}
\end{equation}
\section{Experiments} 
\subsection{Implementation Details}
\noindent \textbf{Datasets. } We conduct comparative experiments and ablation studies on three standard benchmarks to validate the proposed framework, which are  introduced below:
\begin{itemize}
  \item UCSD Ped2 \cite{li2013anomaly} is a medium-scale VAD dataset containing 16 training videos and 12 testing videos. All videos were captured from outdoor scenes with a camera view parallel to the street. The anomalies include biking, skateboarding, and driving on the sidewalk.
  \item CUHK Avenue \cite{lu2013abnormal} is a large-scale dataset with 21 training videos and 16 testing videos. There are 47 abnormal events, such as unusual running and loitering.
  \item ShanghaiTech \cite{luo2017revisit} is a large-scale and challenging dataset for unsupervised VAD, including 330 training videos and 107 testing videos. It contains 130 abnormal events from 13 different scenes.
\end{itemize}

\noindent\textbf{Evaluation metric. }VAD is a regression task aiming to calculate an anomaly score in the range of $[0, 1]$. The score for a normal frame should be close to 0, while that of an abnormal one should be 1.  We calculate true positive rates and false-positive rates under numerous thresholds. The frame-level area under the curve (AUC) of the receiver operation characteristic is used as an evaluation metric. 

\noindent\textbf{Training details. }The STM-AE framework is trained using the Pytorch framework with an Nvidia Geforce RTX 2080Ti GPU. Adam is used as the optimizer with a batch size of 8. The initial learning rate is set to $4\times10^{-4}$ and is decayed by the cosine annealing methods. All video frames are firstly resized to $256\times 256$ pixels. The framework takes four successive frames as input and predicts the next frame. We set $N$ and $k$ to 32 and 8. The trade-off parameters $\lambda_1$, $\lambda_2$ and $\lambda_3$ are set to $0.6$, $0.01$ and $0.05$, respectively.
\subsection{Quantitative Comparison}

\begin{table}[]
  \centering
  \caption{Results of the frame-level AUC comparison. Bolded numbers indicate the best performance, and underlined ones indicate the second-best performance.}
  \label{tab:1}
  \resizebox{.48\textwidth}{!}{
  \begin{tabular}{@{}lllccc@{}}
  \toprule
  \multicolumn{2}{c}{\multirow{2}{*}{\textbf{Type}}} & \multicolumn{1}{l}{\multirow{2}{*}{\textbf{Method}}} & \multicolumn{3}{c}{\textbf{Frame-level AUC (\%)}} \\ \cmidrule(l){4-6} 
  \multicolumn{2}{c}{}    & \multicolumn{1}{c}{}       & UCSD Ped2       & CUHK Avenue    & ShanghaiTech   \\ \midrule
  \multirow{4}{*}{\rotatebox[]{90}{\textbf{Shadow}}} 
 \multirow{4}{*}{\rotatebox[]{90}{}}        
  &   & Kim \textit{et al.} \cite{kim2009observe}            & 69.3   & -     & -     \\
  &   & Lu \textit{et al.} \cite{lu2013abnormal}     & -   &  80.9    & -     \\
  &   & Xu \textit{et al.} \cite{xu2017detecting}     & 90.8   & -     & -     \\
  &   & Tudor \textit{et al.} \cite{tudor2017unmasking}         & 82.2   & 80.6  & -     \\ \midrule
\multirow{13}{*}{\rotatebox[]{90}{\textbf{Deep learning-based}}} 
& \multirow{6}{*}{\rotatebox[]{90}{Single-task}}  
                & Luo \textit{et al.} \cite{luo2017revisit}         & 92.2   & 81.7  & 68.0  \\
        &        & Gong \textit{et al.} \cite{gong2019memorizing}            & 94.1   & 83.3  & 71.2  \\
        &        & Zhang \textit{et al.} \cite{zhang2020normality}       & 95.4   & 86.8  & 73.6  \\
        &        & Park \textit{et al.} \cite{park2020learning}        & 97.0   & 88.5  & 72.8  \\
        &        & Ye \textit{et al.} \cite{ye2019anopcn}           & 96.8   & 86.2  & 73.6  \\
        &        & Liu \textit{et al.} \cite{liu2018future}        & 95.4   & 85.1  & 72.8  \\ \cmidrule(l){2-6} 
& \multirow{5.5}{*}{\rotatebox[]{90}{Two-stream}} 
                & Cai \textit{et al.} \cite{cai2021appearance}          & 96.6   & 86.6  & 73.7  \\
        &        & Chang \textit{et al.} \cite{chang2020clustering}    & 96.5   & 86.0  & 73.3  \\
        &        & Doshi \textit{et al.} \cite{doshi2020continual}      & {\ul 97.8}      & 86.4  & 71.6  \\
        &        & Yu \textit{et al.}  \cite{yu2020cloze}           & 97.3   & {\ul 89.6}     & \textbf{74.6}  \\ \cmidrule(l){3-6} 
        &        & STM-AE (Ours)     & \textbf{98.1}   & \textbf{89.8}  & {\ul 73.8}     \\ \bottomrule
  \end{tabular}}
\end{table}
We quantitatively compared the frame-level AUC of the proposed STM-AE framework with existing unsupervised VAD methods on the UCSD Ped2 \cite{li2013anomaly}, CUHK Avenue \cite{lu2013abnormal} and ShanghaiTech \cite{luo2017revisit} datasets. The results are reported in Table~\ref{tab:1}. The methods involved in the comparison include traditional shallow methods \cite{kim2009observe,lu2013abnormal,xu2017detecting,tudor2017unmasking}, methods with single proxy task (single-task) \cite{luo2017revisit,gong2019memorizing,zhang2020normality,park2020learning,ye2019anopcn,liu2018future} and two-stream structure-based (two-stream) methods \cite{cai2021appearance,chang2020clustering,doshi2020continual,yu2020cloze}. Our STM-AE framework outperforms the state-of-the-art methods on UCSD Ped2 and CUHK Avenue datasets, achieving AUCs of 98.1\% and 89.8\%, respectively, which are significantly higher than shallow and single-task methods. The two-stream methods generally outperform single-based methods, indicating that separate consideration of spatial normality and temporal normality is effective to address unsupervised VAD. Further, our STM-AE framework has explored the correlations between spatial and temporal features. 
  
On the ShanghaiTech dataset, the STM-AE framework achieves the second-best result, slightly lower than \cite{yu2020cloze} by 0.8\%. A possible reason is that our spatial-temporal memories cannot adapt to data from different scenes. Compared to the single-scene of UCSD Ped2 and CUHK Avenue, ShanghaiTech includes videos from 13 different scenes. Additionally, the STM-AE framework takes 0.28 seconds on average to calculate the anomaly scores of a given frame, i.e., the inference speed is around 40 fps, satisfying the needs of real-time detection in real-world applications.
\subsection{Ablation Studies}

\begin{table}[]
  \centering
  \caption{Results of ablation studies. We report frame-AUC on the UCSD Ped2 dataset. Bolded number indicates the best performance, and underlined one indicates the second-best.}
  \label{tab:2}
  \resizebox{.48\textwidth}{!}{
  \begin{tabular}{c|cccc|cccc|c}
  \hline
  \textbf{Model} & $AE_a$  & $\bm{M}_s$   & $AE_m$  & $\bm{M}_t$   & $\mathcal{L}_a$   & $\mathcal{L}_m$ & $\mathcal{L}_r$ & $\mathcal{L}_{adv}$ & \textbf{AUC (\%)} \\ \hline
  1              &  \checkmark &  \checkmark &   &    &  \checkmark &   &  \checkmark &  \checkmark & 96.8              \\
  2              &    &    &  \checkmark &  \checkmark &   &  \checkmark &  \checkmark &  \checkmark & 93.6              \\
  3              &  \checkmark &    &  \checkmark &    &  \checkmark &  \checkmark &  &  \checkmark & 95.5              \\ \hline
  4              &  \checkmark &  \checkmark &  \checkmark &  \checkmark &  \checkmark &  \checkmark &  \checkmark &  \checkmark & \textbf{98.1}     \\ \hline
  5              &  \checkmark &  \checkmark &  \checkmark &  \checkmark &  \checkmark &    &  \checkmark &    & 96.9              \\
  6              &  \checkmark &  \checkmark &  \checkmark &  \checkmark &  \checkmark &  \checkmark &  \checkmark &    & 97.4              \\
  7              &  \checkmark &  \checkmark &  \checkmark &  \checkmark &  \checkmark &    &  \checkmark &  \checkmark & {\ul 97.5}        \\
  8              &  \checkmark &  \checkmark &  \checkmark &  \checkmark &  \checkmark & \checkmark   &   &  \checkmark &  96.3       \\ \hline
  \end{tabular}}
  \vspace{-10pt}
  \end{table}
  To verify the effectiveness of joint normality learning, spatial-temporal memories, and training loss, we conducted ablation studies on the UCSD Ped2 dataset. The results are presented in Table 2, and the discussion is below.

  \textbf{Effectiveness of appearance and motion normality learning. }To demonstrate the advantages of joint learning of appearance and motion normality over single normality learning, we compared the performance of models with different component combinations. The joint normality (model 4) achieves a 1.3\% and 4.5\%  AUC improvement compared to single appearance normality (model 1) and single motion normality (model 2), respectively, demonstrating the effectiveness of the two-stream structure. For unsupervised VAD, the appearance normality seems more critical than the motion normality. Additionally, Compared to model 4, model 3 remove the spatial-temporal memories and directly concatenate the features from $f_a^e$ and $f_m^e$ during the decoding process. The 2.6\% AUC gap demonstrates that the spatial-temporal memories can store prototypical features of normal events and effectively improve the performance of unsupervised VAD tasks. 
  
  \textbf{Effectiveness of training loss. }Models 5-7 explore the effectiveness of training loss, and the results showed that simultaneous constraints on the completeness of appearance and motion information are meaningful. Comparing the performance of model 7 with that of model 4,  we find that the patch discriminator brings a 0.6\% AUC gain, indicating that adversarial learning can improve the ability to fuse and decode appearance and motion features.

\begin{figure}
    \centering
    \includegraphics[width=.46\textwidth]{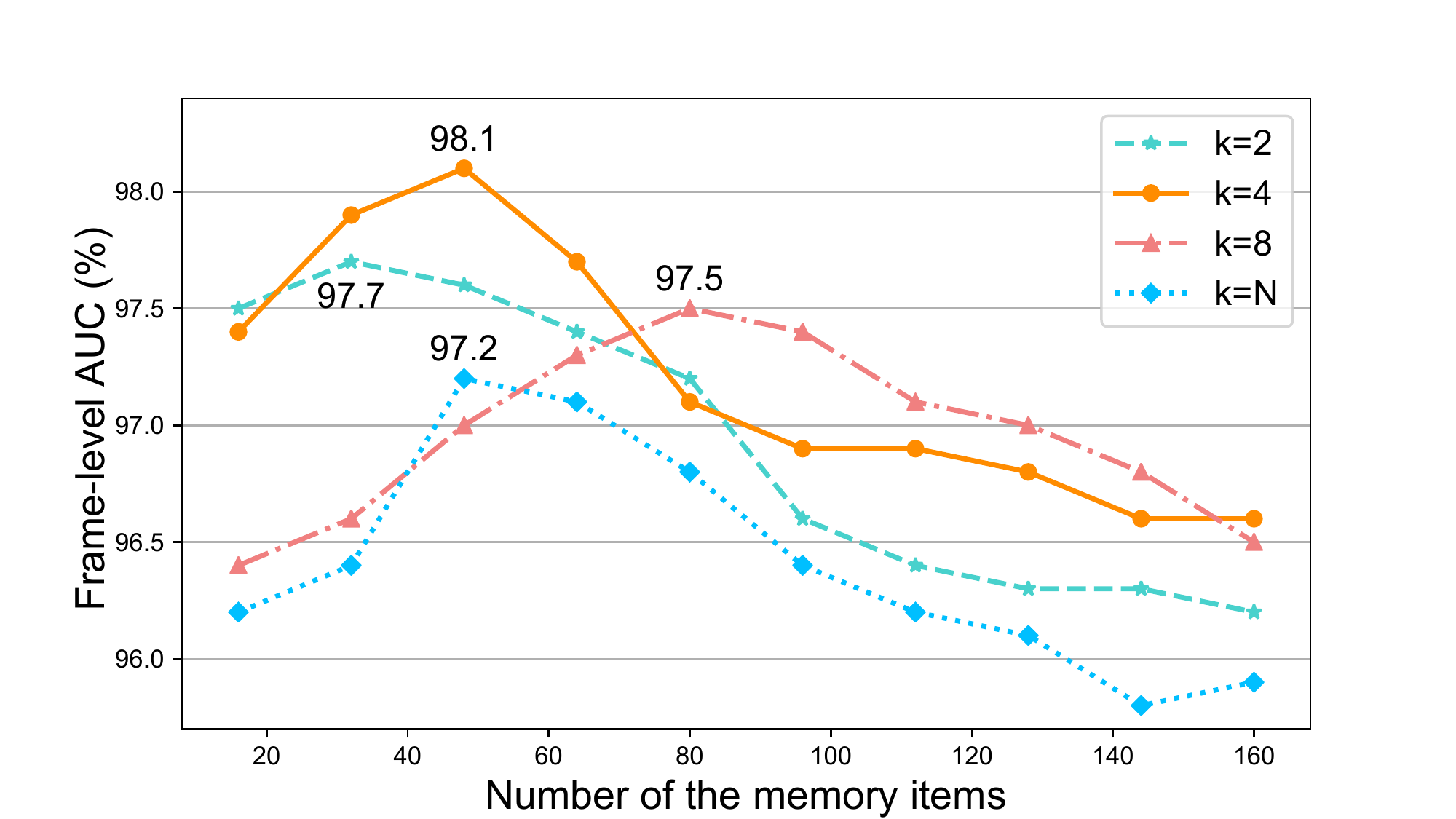}
    \caption{Results of ablation studies on the sensitivity.}
    \label{fig:k1}
\end{figure}

  \textbf{Sensitivity to $N$ and $k$. }In addition, we explore the effect of the number of memory items in the memory pool and the number of selected memory items in the write operations, i.e., the sensitivity of the STM-AE framework to the hyperparameters $N$ and $k$. The results are shown in Figure~\ref{fig:k1}. As the value of $N$ increases, the AUC with different $k$ rises first and then falls. A small memory pool will lose information due to insufficient capacity.  In contrast, a large memory pool will record additional worthless features that cannot represent the prototypical pattern of normal events. The curves with fixed $k$ (2, 4, 8) are generally above the curve with no constraint on  $k$ values (marked by triangles), indicating that updating only top-$k$ relevant memory items instead of all help the memories to learn prototypical spatial and temporal patterns of normal events. Qualitatively, our updating mechanism brings a 0.9\%  AUC gain on the UCSD Ped2 dataset.
\begin{figure}
  \centering
  \includegraphics[width=.45\textwidth]{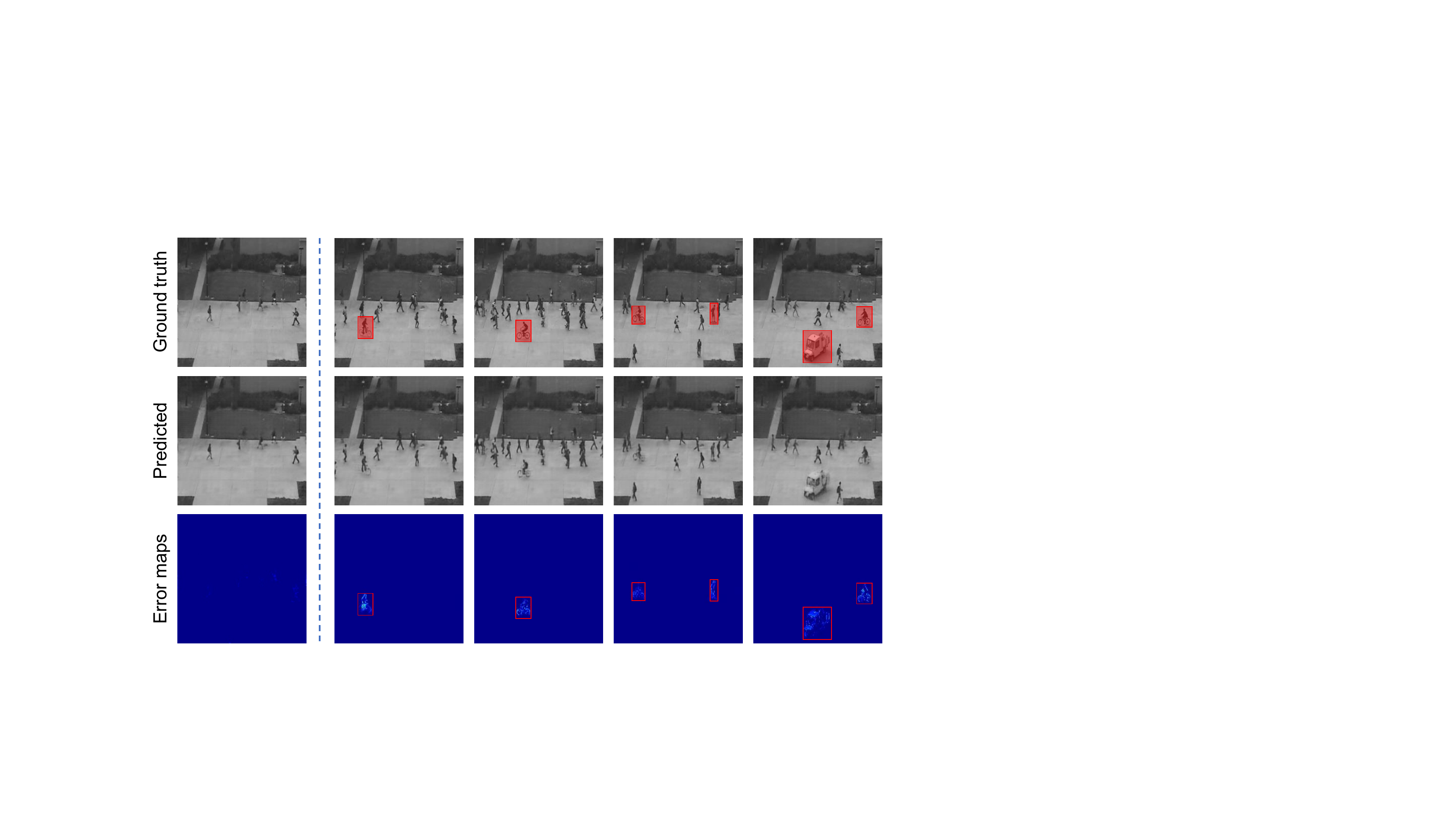}
  \caption{Results of spatial localization of abnormal events.}
  \label{fig:res}
\end{figure}
\subsection{Visual Results}
In Figure~\ref{fig:res}, we show the spatial localization results of the STM-AE framework for abnormal events. The experiments are conducted on the UCSD Ped2 \cite{li2013anomaly} dataset. Rows 1-3 are the ground-truth future frames, predicted frames, and prediction errors. The first column is for normal walking, while columns 2-4 are for abnormal events, including riding, skateboarding, and driving on the sidewalk. The regions marked by red boxes are the locations where abnormal events occur. Comparing columns 1 and 2-4, we find that the STM-AE framework can predict normal events effectively with minor errors, especially for the background parts. For abnormal frames, the prediction results for regions containing abnormal events are significantly worse than for normal motion and background areas, indicating that the STM-AE framework can localize unusual appearance and motion.
\section{Conclusion}
In this paper, we address unsupervised VAD by considering appearance anomalies and motion anomalies separately. The proposed STM-AE framework learns prototypical appearance and spatial patterns separately and records them in spatial-temporal memories. And the intrinsic correlation between appearance and motion normality is investigated via adversarial learning. Experimental results demonstrate the effectiveness of joint normality learning, and the STM-AE framework outperforms the state-of-the-art methods. For the problem of performance degradation in processing multi-scene videos, we will explore hierarchical memories to improve the adaptability to multi-scene data in future work.
\bibliographystyle{IEEEbib}
\bibliography{mylib}
\end{document}